%% file: blk_hyper.tex
\documentclass[12pt]{article}
\usepackage{fullpage}
\usepackage[english]{babel}
\usepackage{amsmath}
\usepackage{bm} 
\usepackage{multirow}
\usepackage{multicol} 
\usepackage{appendix} 
\usepackage[neverdecrease]{paralist} 
\usepackage{graphicx} 
\usepackage{hyperref} 
\usepackage[noabbrev]{cleveref}
\usepackage{pgfplots}
\usepackage{enumitem}

\usepackage{fancyhdr} 
\usepackage{algpseudocode, algorithm, algorithmicx} 
\usepackage{wrapfig}

\input{header.tex}

\usepackage[style=ieee]{biblatex}
\newcommand{\reals}{{\mbox{\bf R}}}
\newcommand{\ie}{{\it i.e.}}

\title{
	{\bfseries A Light-Weight Multi-Objective Asynchronous Hyper-Parameter
Optimizer}
}

\author{
Gabriel Maher \thanks{email: gabriel.maher@blackrock.com} \and
Stephen Boyd \and Mykel Kochenderfer \and Cristian Matache
\and Alex Ulitsky \and Slava Yukhymuk \and Leonid Kopman
}

\date{BlackRock AI Lab, \today
}

\begin{document}

\maketitle

\abstract{
We describe a light-weight yet performant system for hyper-parameter optimization
that approximately minimizes an overall scalar cost function that is obtained by combining
multiple performance objectives using a target-priority-limit scalarizer.
It also supports a trade-off mode, where the goal is to find an appropriate
trade-off among objectives by interacting with the user.
We focus on the common scenario where there are on the order of tens of hyper-parameters,
each with various attributes such as a range of continuous values, or a finite list of values,
and whether it should be treated on a linear or logarithmic scale.
The system supports multiple asynchronous simulations and is robust to
simulation stragglers and failures.
}

\section{Hyper-parameters, objectives, and cost}

\paragraph{Hyper-parameters.}
We let $x \in \mathcal X \subset \reals^n$ denote the vector of
hyper-parameters that we will choose from the set of
valid choices $\mathcal X$, to be discussed in \cref{s-attributes}).
We refer to $x_i$ as the $i$th hyper-parameter.
While the algorithm we describe will work in principle for
any value of $n$, our focus is on the case where $n$ is modest (e.g., $< 10$).

\paragraph{Objectives.}
We have $k$ objectives,
denoted $f_i: \mathcal X \to \reals$, that we use
to evaluate a choice of $x$.
Each of these objectives has a sense, which means that we either want the objective to
be large (\ie, to maximize it) or small (\ie, to minimize it).
We refer to these as maximization and minimization objectives, respectively.
We use $f$ to denote the vector-valued objective, $f(x) = (f_1(x), \ldots, f_k(x))$.

\paragraph{Overall cost.}
To compare different choices of hyper-parameters, we
scalarize the objectives to obtain an overall cost which we denote
$F: \mathcal X \to \reals \cup\{\infty\}$, which we assume we wish to
minimize, with $F(x) = \infty$ meaning an unacceptable value of $x$.
We express this as
\[
F(x) = \phi(f(x)) = \phi(f_1(x), \ldots, f_k(x)),
\]
where $\phi:\reals^k \to \reals \cup \{\infty\}$ is the scalarization function,
which is monotone nondecreasing in each argument corresponding to
a minimization objective,
and monotone nonincreasing in each argument corresponding
to a maximization objective.
(Roughly speaking, this means that $F$ decreases when a minimization
objective decreases or when a maximization objective increases.)

A very simple scalarizer is $\phi(u) = w^T u$, where $w\in \reals^n$ is a
set of weights, with $w_i \geq 0$ for minimization objectives and
$w_i \leq 0$ for maximization objectives.  In this case, $F(x)$ is
a traditional weighted sum of the objectives.

\paragraph{Target-priority-limit scalarizer.}
The target-priority-limit scalarizer is a bit more complex than the weighted-sum
scalarizer, while still being expressive and interpretable.
It is separable, \ie, it has the form
\[
\phi(u) = \sum_{i=1}^k \phi_i(u_i),
\]
where $\phi_i: \reals \to \reals \cup \{ \infty\}$.
Each $\phi_i$ is characterized by three numbers: a \emph{target}
value $T_i$,
a \emph{priority} $P_i>0$, and a \emph{limit} $L_i$.
For an objective that we want to minimize, we require that $T_i \leq L_i$,
and have
\[
\phi_i(u_i) = \left\{ \begin{array}{ll} 0 & u_i \leq T_i \\
P_i\frac{u_i-T_i}{L_i-T_i} & T_i \leq u_i \leq L_i \\
\infty & u_i > L_i.
\end{array}\right.
\]
We note that $\phi_i(u_i) \geq 0$, taking its smallest value $0$
only when the objective is less than or equal to the target value.
When the objective exceeds the target value but not the limit,
$\phi_i(u_i)$ is proportional to the fractional distance between these
two values, with slope $P_i$.  If $u_i$ exceeds the limit, we have
$\phi_i(u_i)=\infty$.
In other words, the target is the value of the objective below
which we are equally satisfied, and the limit specifies the maximum allowed
value.  The priority sets the slope in between.

For an objective we wish to maximize, we require $L_i \leq T_i$ and
reverse the function as
\[
\phi_i(u_i) = \left\{ \begin{array}{ll} 0 & u_i \geq T_i \\
P_i\frac{T_i-u_i}{T_i-L_i} & L_i \leq u_i \leq T_i \\
\infty & u_i < L_i.
\end{array}\right.
\]

\begin{figure}
    \begin{center}
        \input{figures/obj.tex}
    \end{center}
    \caption{An example of the target-priority-limit scalarizer,
with one minimization objective and one maximization objective.
The overall cost is $\phi_1(f_1(x))+\phi_2(f_2(x))$.
\label{fig:obj}}
\end{figure}
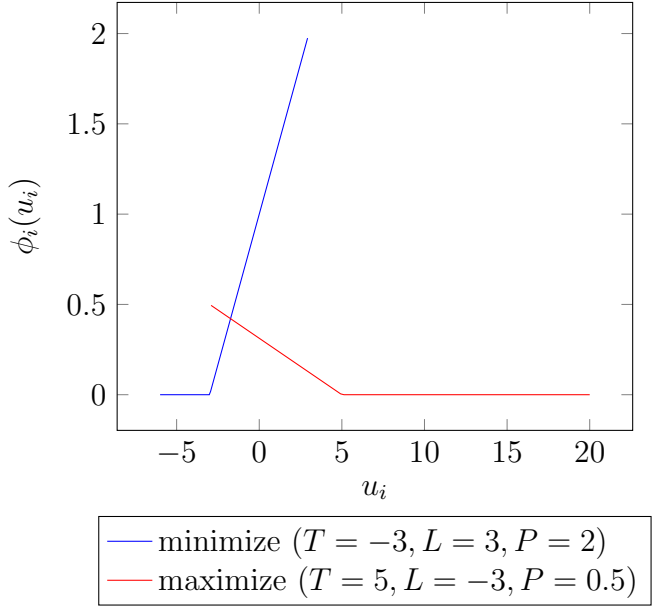

\Cref{fig:obj} shows two examples of $\phi_i$, one for a minimization
objective and one for a maximization objective.

\section{Hyper-parameter optimizer}

\paragraph{Simulation.}
We assume that the objectives $f_1, \ldots, f_k$
are given as black-box oracles that we
can evaluate.  We cannot evaluate
gradients of $f_i$; in fact, $f_i$ is not required to be differentiable.
We refer to evaluating the objectives $f(x) = (f_1(x), \ldots, f_k(x))$
as a \emph{simulation}, which can involve substantial computation.
We assume that simulations can be carried out by a set of workers or agents.

\paragraph{Hyper-parameter optimizer.}
Our goal is to find a good value of the hyper-parameters, \ie, to find $x \in
\mathcal X$ that approximately minimizes $F(x)$, using only simulations.
Thus we approximately solve the hyper-parameter optimization problem
\[
\begin{array}{ll}
\mbox{minimize} & F(x) \\
\mbox{subject to} & x \in \mathcal X,
\end{array}
\]
with variable $x$, using only evaluations of $F$, \ie,
simulations.

\paragraph{Trade-off mode.}
As a variation on scalarization of the objectives, we can also
search for Pareto optimal points with respect to two or three of the objectives.
Our goal is to explore the trade-off curve (for two objectives) or
trade-off surface (three objectives).
This mode is described in \cref{s-trade-off-mode}.
Until then, we consider the scalarized hyper-parameter problem
described above.

\paragraph{Hyper-parameter optimizer state.}
Our hyper-parameter optimizer is a centralized entity
that keeps track of previous simulations, \ie, a list
\[
(x^i, f(x^i)), \quad i=1, \ldots, K,
\]
where $K$ is the number of simulations carried out so far,
$x^i\in \mathcal X$, and the superscript $i$ indexes the simulations.
This list is the state of the optimizer.
For each of these simulations we can evaluate the
overall cost $F(x^i)=\phi(f(x^i))$,
and sort the list of previous simulations from
best (\ie, smallest $F(x^i)$) to worst (\ie, largest $F(x^i)$).
The best $r$ simulations can be displayed as a leader-board.

\paragraph{Interaction with workers.}
The optimizer interacts with multiple workers (such as
CPUs, processes, or threads) that carry out simulations asynchronously.
When a worker is available to carry out a simulation, it
queries the optimizer to obtain a recommended value of $x$ to simulate.
When a worker finishes a simulation, it reports
values of the $n$ hyper-parameters and the $k$ objectives found, which is added
to the list of previous simulations.
We do not assume that the workers are reliable; that is, we assume they
can take a long time to evaluate the objectives for a given value of $x$,
or even fail, \ie, never report back.

\section{Parameter attributes}\label{s-attributes}

Each hyper-parameter can have a number of attributes,
which we describe here.
(Not all combinations of these attributes make sense.)
These collectively give the set of allowed values $\mathcal X$.

\paragraph{Range.}
Each hyper-parameter has a minimum and maximum allowed value,
denoted $l_i$ and $u_i$, with $l_i < u_i$.

\paragraph{Linear or logarithmic scale.}
Each hyper-parameter has either a linear or logarithmic scale.
If it is logarithmic, we require that $l_i>0$.
The choice will affect how values are treated internally, and the meaning
of other attributes such as gridded values, described below.

\paragraph{Grid and finite set of values.}
We can require that a hyper-parameter take on only a specified
finite set of values.  For example, we can require that $x_i \in \{0,1\}$,
\ie, that the $i$th hyper-parameter is Boolean.
A common special case is a grid of $N$ values, given by
\[
x_i \in \{ l_i, l_i+\delta, \ldots, l_i + (N-1) \delta\}, \quad
\delta = \frac{u_i-l_i}{N-1},
\]
when $x_i$ has linear scale, or
\[
x_i \in
\{l_i, \gamma l_i, \ldots,  \gamma^{N-1}l_i\}, \quad
\gamma = (u_i/l_i)^{1/(N-1)},
\]
when $x_i$ has logarithmic scale.
Another common specification of a finite set of values is that $x_i$
should be an integer, \ie,
\[
x_i \in \{ \lceil l_i \rceil,
\lceil l_i \rceil + 1, \ldots, \lfloor u_i \rfloor \}.
\]

%

\section{Method}

\subsection{Standarized hyper-parameters}

\paragraph{Standardizing the hyper-parameters.}
We first transform each $x_i$ to $z_i$, a standardized hyper-parameter
value in $[0,1]$.
For a linear scale hyper-parameter, we use
\[
z_i = \frac{x_i-l_i}{u_i-l_i},
\]
and for a logarithmic scale hyper-parameter,
\[
z_i =
\frac{\log x_i-\log l_i}{\log u_i- \log l_i}.
\]
The standardized hyper-parameter $z$ lies in $[0,1]^n$.
We let $\mathcal Z$ correspond to $\mathcal X$ after standardization.  Thus $z \in \mathcal Z$ means
that the corresponding $x$ satisfies $x \in \mathcal X$.

\paragraph{Un-standardizing.}
Our method will suggest values of $z \in \reals^n$, which are transformed back to $x$
by first projecting onto $\mathcal Z$, and then inverting the standardization mappings.
Given the suggestion $z \in [0,1]^n$, we first form
\[
\tilde z = \Pi_{\mathcal Z} (z),
\]
where $\Pi_{\mathcal Z}$ is projection onto $\mathcal Z$.
This projection involves clipping the entries $z_i$ to $[0,1]$
and then rounding to the nearest valid value of $z_i$.
We then un-standardize using
\[
x_i = l_i + z_i(u_i-l_i)
\]
for linear scale hyper-parameters and
\[
x_i = \exp \left( \log  l_i + z_i(\log u_i- \log l_i) \right)
\]
for logarithmic scale hyper-parameters.

\subsection{Sampling and exploration strategy}
When a worker requests a hyper-parameter value to simulate,
the optimizer generates $z$ randomly from some distribution,
and then un-standardizes it as described above to obtain a valid suggestion
$x \in \mathcal X$.
We now explain how we represent and update this distribution.

\paragraph{Initial exploration phase.}
We let $K$ denote the total number of simulations carried out so far.
For $K$ less than a given threshold,
we simply sample $z$ from a uniform distribution on $[0,1]^n$.
Our default threshold is
\[
\min\{ \lfloor S/5\rfloor, 50 + 2n\},
\]
where $S$ is the intended number of simulations.
Once $K$ equals the threshold, we switch to a more
sophisticated distribution that depends on the state of the optimizer, \ie,
the table of previous simulations.

\paragraph{Elite points.}
We work with the elite points, which is the set of the $r$ best values we have
found so far, judged in terms of overall cost.
We take $r$ to be some fraction $\eta$ (e.g., $0.2$) of $K$, the total number of
simulations we have carried out.  In other words, the elite points
are the ones with cost at or below the $\eta$-quantile among the
simulations.

\paragraph{Statistical model of elite points.}
We fit a Gaussian mixture model of the standardized versions of the
elite points.
This statistical model is updated whenever the elite set changes,
which can occur when new simulation results
arrive, or when the scalarization changes.
When a worker requests a new hyper-parameter value,
we sample from this distribution, project onto $\mathcal Z$, and then un-standardize.

\paragraph{Low discrepancy sample points.}
Instead of sampling from a uniform distribution for the initial phase,
we instead use Sobol points \cite{Sobol1967}, which are more space filling than random samples.
For the Gaussian mixture model, we may also sample from points derived from
a Sobol sequence. To generate these Gauss-Sobol points, we find a Sobol sequence in
$\reals^2$, and then transform these in the standard way to obtain points
in $\reals^2$ that would have had an $\mathcal N(0,I)$ distribution had
the points been uniform.
These, in turn, are transformed by an appropriate affine mapping to obtain
the desired mean and covariance.

%

\section{Trade-off mode}\label{s-trade-off-mode}

This tool supports multi-objective optimization in a trade-off mode.
In this case, we do not scalarize the objectives,
but instead indicate two or three for which we seek Pareto optimal points.
For all other objectives we can maintain limits, \ie, we still consider only
choices for which all metrics are within the limits, but we do not set priorities or target values.
Our goal is to explore the trade-off curve (for two objectives) or
trade-off surface (for three objectives).

To handle this trade-off exploration, we only need to change how we
characterize previous simulations as elite or not.
Similar to \citeauthor{Deb2002} \cite{Deb2002}, we define the level~1 simulations as
those that are Pareto optimal with respect to the given two or three objectives.
Then we define the level~2 simulations to be those that are Pareto optimal
after the level~1 simulations are removed. We continue this for other levels
until all past simulations have been assigned a level.
To choose $r$ previous simulations, we first consider the level~1 simulations.
If $r$ is smaller than or equal to the number of level~1 simulations, we
randomly choose $r$ of them.  Otherwise we label all level~1 simulations as
elite and move to level~2 simulations to find more elite points, repeating
this process.
In other words, we sort the simulations according to their Pareto level
instead of a scalarized overall objective value; otherwise, the
search algorithm is the same.

\section{The HOLA software package}

We implemented the method described above in the Python package
HOLA (Hyper-parameter Optimizer: Light-weight and Asynchronous).
We describe HOLA here.

\paragraph{Objective and scalarizer configuration.}
The objective names, targets, limits and priorities are specified as a key-value dictionary, for example
\begin{verbatim}
objectives_config = {
    "accuracy":{
        "target": 1.0,
        "limit": 0.0,
        "priority": 2.0
    },
    "abs_error":{
        "target": 0,
        "limit": 1000,
        "priority": 0.5
    }
}
\end{verbatim}

\paragraph{Hyper-parameter search space configuration.}
The search-space for each hyper-parameter is specified with a few options supplied to our software as a simple key-value dictionary.
First, users must specify a \verb|min| and \verb|max| value.
Thereafter, the scale can be set to either linear or logarithmic and defaults to linear if no choice is made.
Users can also elect to sample from an equally spaced grid of points between \verb|min| and \verb|max|.
Note that the grid will be equally spaced on a log scale if the scale is set to logarithmic.
Finally hyper-parameter values can be sampled from a set of discrete values as well.
An example hyper-parameter search-space configuration could look like
\begin{verbatim}
params_config = {
    "n_estimators":{
        "min": 10,
        "max": 1000,
        "param_type": "int",
        "scale": "log"
        "grid": 10
    },
    "max_depth":{
        "values": [1, 3, 5, 7]
    },
    "learning_rate":{
        "min": 1e-4,
        "max": 1.0,
        "scale": "log"
    },
    "subsample":{
        "min": 0.2,
        "max": 1.0
    }
}
\end{verbatim}

\paragraph{Running hyper-parameter optimization in local mode.}
HOLA exposes a \verb|tune| function that provides a straightforward interface to run hyper-parameter optimization sessions
on a single machine (but still utilizing all available processors).
Users simply pass the objectives, the configuration dictionaries, as well as a stopping criterion called \verb|num_runs|
which corresponds to the number of simulations made until the best hyper-parameters are reported.
Calling
\begin{center}\verb|tune(func, params_config, objectives_config, num_runs=100, n_jobs=2)|\end{center}
spawns \verb|n_jobs| processes that request hyper-parameter suggestions from HOLA and execute \verb|func| until
\verb|num_runs| simulations are complete.
If \verb|n_jobs=-1|, then HOLA will detect the number of available processors and spawn that many processes.

Users connect their simulation to HOLA using the \verb|func| argument.
The inputs to \verb|func| are expected to be the same as the keys in the hyper-parameter search-space configuration dictionary.
\verb|func| is expected to return a key-value dictionary with objective names and corresponding objective values as keys and values respectively.
In other words, \verb|func| should accept the hyper-parameters as input, run a simulation and return a key-value dictionary of the resulting objective values.

The \verb|tune| function returns an underlying \verb|Tuner| object which is responsible for running the optimizer and maintaining the state.
While running simulations, the \verb|Tuner| object will simultaneously sort the leader-board of simulation results and provide optimized hyper-parameter suggestions.
Calling \verb|Tuner.get_best_params()| will return the current best hyper-parameter values in the leader-board.
Similarly, \verb|Tuner.get_best_scores()| will return the best objective and scalarized score values.

\paragraph{Saving and restoring hyper-parameter optimization sessions.}
Our package stores completed simulation results in a leader-board that can be saved to a dataframe at any time.
A hyper-parameter optimization session can be restored by loading a saved leader-board dataframe, supplying an objectives configuration dictionary, hyper-parameter search-space configuration dictionary, and the simulation function.
It is thus straightforward to save and restore hyper-parameter optimization sessions using our package.

\paragraph{HOLA hyper-parameter optimization server.}
We provide a server implementation of the HOLA package such that the workers can be implemented and run completely independently, for example across multiple machines or in a different programming language.
The HOLA server exposes a simple HTTP interface through which workers can request hyper-parameter samples and report simulation results.
Note that we assume a single server instance will be responsible for a single family of simulation results, that is all simulation results reported by workers correspond to the same hyper-parameter and objectives configuration.
Futhermore the leaderboard of hyper-parameters and simulation results can be viewed in a web browser in real-time.
The server is launched by specifiying a directory containing a JSON hyper-parameter and objective configuration file named \verb|hola_params.json| and \verb|hola_objectives.json| respectively.
Optionally, if the directory contains a saved leaderboard csv file named \verb|hola_results.csv|, the leaderboard will be restored and the server will resume from that point on.
By default the server will run on the \verb|localhost| address on port 8675, but an alternative port can be specified as a command-line argument.

We expose the following routes through the HOLA server for workers to request hyper-parameter samples and report simulations results

\begin{itemize}
\item \verb|/|
	\begin{itemize}
		\item \verb|GET|: Will return a HTML page with the current leaderboard results, viewable in a browser.
	\end{itemize}
\item \verb|/report_request|
	\begin{itemize}
		\item \verb|GET|: Will return a hyper-parameter sample as a JSON response with the keys being the hyper-parameter names and the values being their sampled values.
		\item \verb|POST|
			\begin{itemize}
				\item \verb|request|: The request can be empty or optionally a JSON dictionary with keys \verb|params| and \verb|objectives|. The \verb|params| should be a dictionary of hyper-parameter names and values. The \verb|objectives| should be a dictionary of objective names and values.
				\item \verb|response|: If the request is empty, will simply return a JSON hyper-parameter sample dictionary. If the request is non-empty, the hyper-parameter sample and simulation result will be recorded in the leaderboard and a JSON response with a new hyper-parameter sample will be returned.
			\end{itemize}
	\end{itemize}
\item \verb|/param|
\begin{itemize}
	\item \verb|GET|: Will return a key-value dictionary of the best hyperparameters seen so far.
\end{itemize}
\item \verb|/experiment|
	\begin{itemize}
		\item \verb|GET|: Will return a JSON response dictionary with keys \verb|params| and \verb|objectives| containing the hyper-parameter and objective configuration dictionaries, respectively.
	\end{itemize}
\end{itemize}

\section{Prior work}
A simple approach to hyper-parameter optimization known as full-factorial evaluation is to discretize the various parameters values and try out all possible combinations of discrete values, corresponding to points on a grid.
Alternatively hyper-parameter values could also be sampled from the hyper-parameter search space uniformly at random and evaluated \cite{Bergstra2012}.
Because the gradient of the simulation results with respect to the hyper-parameters is typically unavailable, gradient-free optimization methods can also be used for hyper-parameter optimization \cite{Kochenderfer2019}.
Search and exploration methods using Gaussian Processes to model the hyper-parameter distribution have also been applied \cite{Bergstra2011}.
Our method is a modified form of the cross-entropy method \cite{Rubinstein2004}.
There are many open-source software packages dedicated to hyper-parameter optimization such as Spearmint \cite{Snoek2012}, Optuna \cite{Akiba2019}, Tune \cite{Liaw2018}, HyperOpt \cite{Bergstra2013}, and SigOpt \cite{Dewancker2016}.

\clearpage

\printbibliography

\clearpage

\appendix

\section{Algorithm evaluation}\label{sec:algorithm-evaluation}
    This paper introduces not only a general multi-objective optimization framework that could work with several
    optimization algorithms but also a new such algorithm.
    To understand how our HOLA algorithm performs relative to other well-known algorithms, we studied the quality of the
    best values found on different benchmarks problems with different numbers of simulations.

    \subsection{Optimization algorithms}\label{subsec:optimization-algorithms}
    We compared our algorithm with a few other standard algorithms.
    \begin{enumerate}[noitemsep]
        \item \emph{Random Search.} The algorithm suggests new hyper-parameters by sampling random values for each of them.
        \item \emph{Sobol Search.} Instead of sampling randomly, this algorithm uses a Sobol sequence.
        \item \emph{Random Search X2.} This algorithm is the same as Random Search but with twice the budget (i.e., twice as many iterations).
        \item \emph{Iterative Grid Refinement (IGR).} This algorithm starts with a grid search with loosely spaced lattices, finds the best
        parameters, and repeats the grid search in smaller subspaces around the best parameters found so far.
        It tightens the spacing between lattices in these subspaces until a stopping criterion is met.
        \item \emph{Nelder Mead.} A method that incrementally updates a set of points that defines a simplex according to a set of rules \cite{Nelder1965}.
        \item \emph{Hooke Jeeves.} A method that greedily explores points some step size along the various coordinate axes. It reduces the step size if an improvement around a point cannot be found \cite{Hooke1961}.
        \item \emph{Tree-structured Parzen Estimators (TPE).} A Bayesian optimization technique \cite{Bergstra2011}.
        \item \emph{HOLA.} This is our modified version of the cross-entropy method with Gaussian mixture models.
        \item \emph{Genetic Algorithms (GA).} This algorithm evolves a population of points using crossover and mutation operations \cite{Goldberg1989}.
        \item \emph{Particle Swarm Optimization (PSO).} This algorithm moves particles corresponding to points in the design space using various forces that aim to bring the particles close to a global optimum \cite{Kennedy2001}.
    \end{enumerate}
    We used the \texttt{hyperopt} library for TPE, Random Search, and Random Search X2. For Hooke Jeeves, Nelder Mead, GA,
    and PSO, we used the \texttt{pymoo} library. For HOLA, IGR, and Sobol Search, we used our own implementation.

    When running an optimizer, the user typically specifies when the algorithm will terminate.
    In our experiments, we used the number of iterations as our stopping criterion.
    Most of these algorithms are ``informed'' in the sense that they use previous values to make better informed parameter suggestions in the next iteration.

    \subsection{Benchmarks}\label{subsec:benchmarks}
    Here we describe what functions we used for benchmarking our optimization algorithms.
    We ranged the dimensionality of the input domains of these functions between 1D and 7D .

    In terms of real-world scenarios, we maximized the R-squared (by minimizing its opposite) of
    \texttt{GradientBoostingRegressor} from \texttt{scikit-learn} on the built-in \texttt{scikit-learn}
    Diabetes Regression dataset with 4 hyper-parameters: number of estimators, maximum depth, learning rate and
    subsample.

    Moreover, we experimented with several closed-form hard-to-optimize functions: Ackley (2D, 5D, 7D), Branin (2D),
    Bukin 6 (2D), Cross In Tray (2D), Drop Wave (2D), Egg Holder (2D), Forrester (1D), Holder Table (2D), Levy 13 (2D),
    Rastrigin (2D, 5D, 7D), Schwefel (2D, 5D, 7D) and Six Hump Camel (2D) .

    The difficulty in optimizing a function is determined by its modality (unimodal, bimodal, multimodal),
    smoothness and dimensionality.
    For the functions that support an arbitrary number of dimensions (e.g., Rastrigin), the higher the dimensionality
    the harder the problem is to optimize.
    Also, the smoother the function, the easier it should be for an informed optimizer to optimize it.
    Most of the functions we chose are multimodal, and some are maybe quite exaggerated for what would occur in real-world,
    optimization problem, but we wanted to see how far HOLA can go by presenting it challenging problems.
    To balance this, some chosen functions like Branin, Forrester, or Six Hump Camel have fewer local minima.

    \subsection{Experiments and Results}\label{subsec:experiments}
    We ran many times each of the enumerated algorithms on the mentioned benchmarks ranging the number of iterations.

    \subsubsection{Real-world example}
\begin{figure}
        \includegraphics[width=\linewidth]{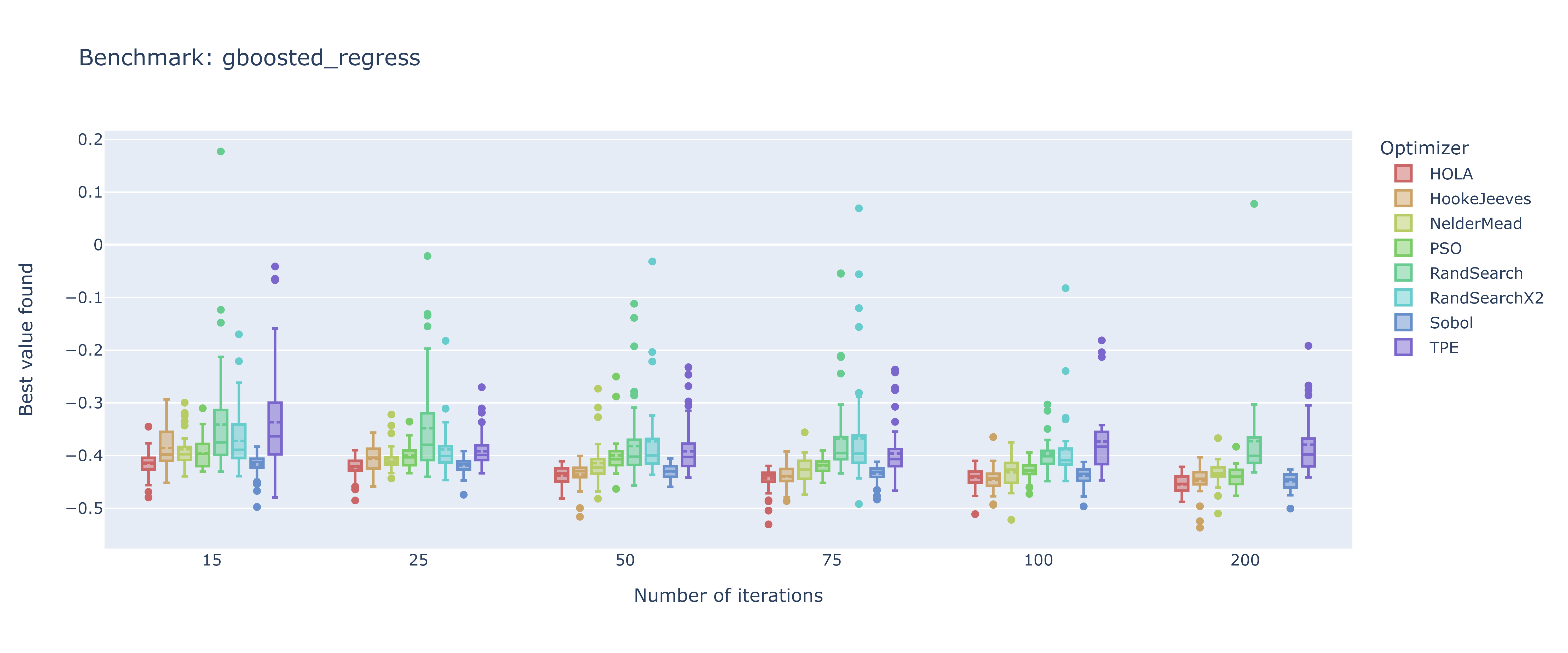}
        \caption{Best values found for Gradient Boosted Regressor on the Diabetes dataset}
        \label{fig:gboosted-box-plot}
    \end{figure}

    For the gradient boosted regressors, we ran at least 50 times each algorithm at 25, 50, 75, 100 and 200 iterations.
    \Cref{fig:gboosted-box-plot} shows the distribution of the best values found by each optimizer at the different
    numbers of iterations in a box plot.
    The median is shown as a solid line while and the mean as a dotted one.
    Since this was implemented as a minimization problem, the lower is better.
    The HOLA algorithm shows that despite its simplicity it is very competitive when compared to the other algorithms.
    Note, that it performs significantly better than TPE which is a method that belongs to the same Bayesian class.

    \subsubsection{Generalization on Synthetic Benchmarks}
    With this encouraging real-world results, we ran further experiments to understand how well HOLA generalizes.
    Therefore, we turned to the synthetic hard-to-optimize functions described in \cref{subsec:benchmarks}.

    Since the performance of algorithms is directly proportional with the number of iterations,
    for the smaller numbers of iterations (25, 50 and 75), we used "easier" benchmarks: maximum 2 dimensional.
    That is: Ackley (2D), Branin (2D), Bukin 6 (2D), Cross In Tray (2D), Drop Wave (2D), Egg Holder (2D),
    Forrester (1D), Holder Table (2D), Levy 13 (2D), Rastrigin (2D), Schwefel (2D), Six Hump Camel (2D) .
    For higher number of iterations, we used "harder" benchmarks: also included 5 and 5 dimensional functions and
    eliminated the ``easier'' functions where all optimizers were optimizing nearly equally well: Forrester, Branin,
    Six Hump Camel, Cross In Tray.
    So, for the higher number of iterations we are left with Ackley (2D, 5D, 7D), Bukin 6 (2D), Drop Wave (2D),
    Egg Holder (2D), Holder Table (2D), Levy 13 (2D), Rastrigin (2D, 5D, 7D), Schwefel (2D, 5D, 7D).

    Again we ran each optimizer on each benchmark for each number of iterations at least 50 times.
    In total, we collected ran 10 optimization algorithms which tried 29668300 combinations of hyper-parameters on 18
    multimodal functions (i.e., benchmarks).

    To compare across benchmarks we min-max normalized these values and computed, first the mean per
    number of iterations, optimizer and benchmark, and then we took again the mean of these means per number of
    iterations and optimizer.
    That is, we can observe the performance of the optimizers for each number of iterations, across all benchmarks by
    plotting this "mean of means of normalized best values found by simulations".

    \begin{figure}
        \centering
        \includegraphics[width=\linewidth]{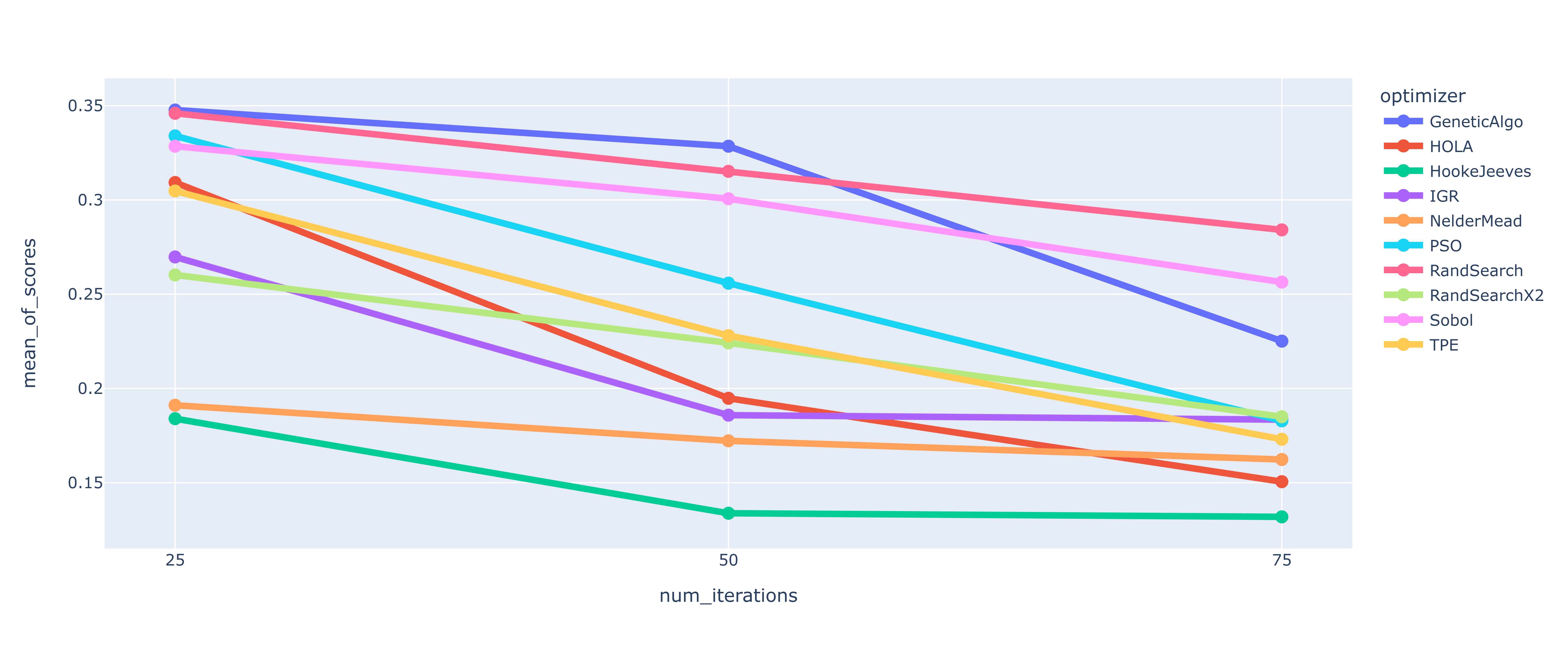}
        \caption{Average performance per optimizer across hard-to-optimize functions}
        \label{fig:average-performance-small-iter}
    \end{figure}

    \begin{figure}
        \centering
        \includegraphics[width=\linewidth]{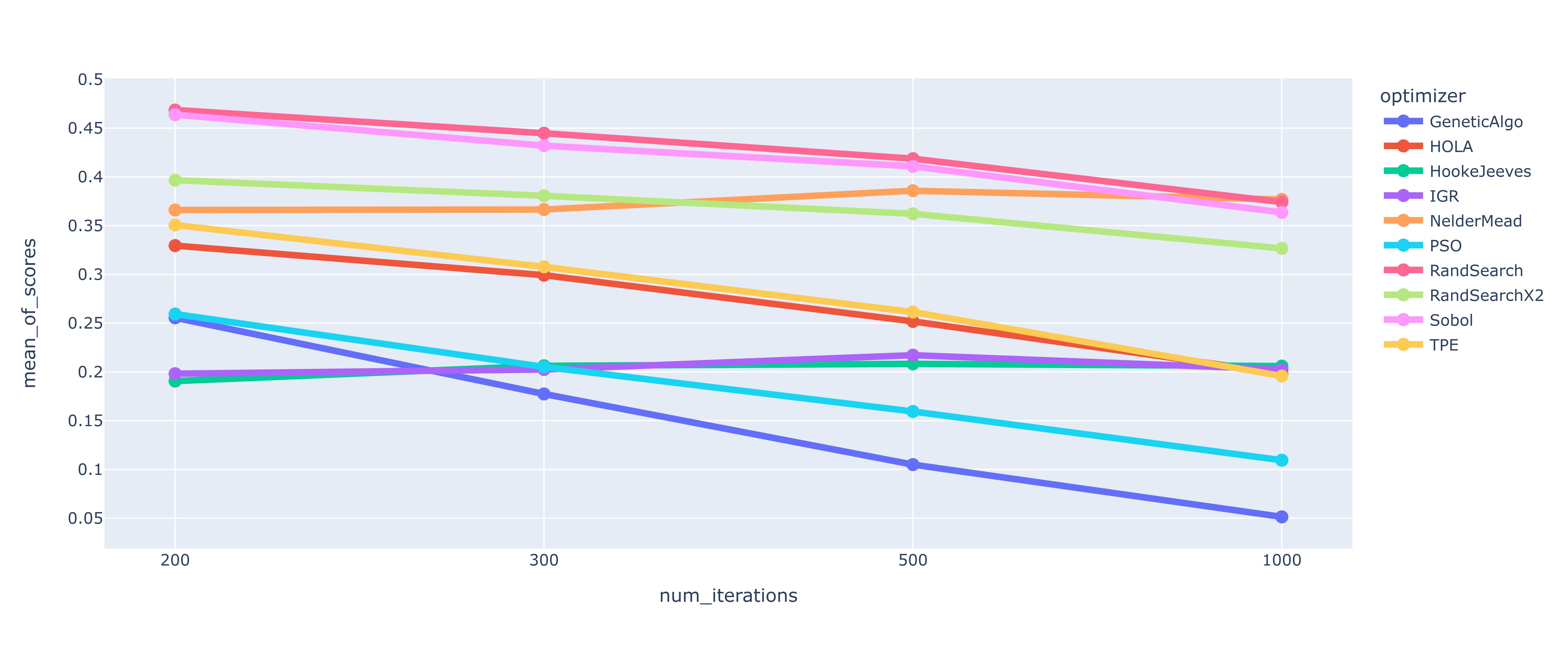}
        \caption{Average performance per optimizer across hard-to-optimize functions}
        \label{fig:average-performance-large-iter}
    \end{figure}

        In \cref{fig:average-performance-small-iter,fig:average-performance-large-iter}, we show these means.
    Since all problems are minizimations lower is better.
    These plots reinforce the existing literature as:
    \begin{itemize}[noitemsep]
        \item The more iterations the better an algorithm performs.
        \item Genetic Algorithms and PSO perform better at a high number of iterations.
        \item Random Search X2 is better than Sobol Search which in turn is better than Random Search.
    \end{itemize}
    Based on this further analysis, we can conclude that HOLA is performant in a more general context of multimodal
    hard-to-optimize functions.
    Random Search X2 is so cheap in terms of resources and simple to implement, but HOLA significantly
    outperforms it with half the resources.
    Moreover, HOLA dominates the popular TPE algorithm (hyperopt), the latter only catching up with HOLA at 1000 iterations.
    When compared to the other algorithms, HOLA is quite simple but nonetheless powerful.
    Similarly, the Iterative Grid Refinement (IGR) algorithm is also very simple and quite performant.

\end{document}

%% file: header.tex
\usepackage{amsmath} 
\usepackage{amsthm} 
\usepackage{amssymb}	
\usepackage{graphicx} 
\usepackage{multicol} 

\let\baraccent=\= 
\renewcommand{\=}[1]{\stackrel{#1}{=}} 

\theoremstyle{definition}

\theoremstyle{remark}


%% file: figures/obj.tex
\begin{tikzpicture}[]
\begin{axis}[
  ylabel = {$\phi_i(u_i)$},
  xlabel = {$u_i$},
  legend cell align={left},legend style={at={(0.5,-0.2)},anchor=north}
]

\addplot+[
  mark = {none}
] coordinates {
  (-6.0, 0.0)
  (-5.919597989949748, 0.0)
  (-5.839195979899498, 0.0)
  (-5.758793969849246, 0.0)
  (-5.678391959798995, 0.0)
  (-5.597989949748744, 0.0)
  (-5.517587939698492, 0.0)
  (-5.437185929648241, 0.0)
  (-5.35678391959799, 0.0)
  (-5.276381909547739, 0.0)
  (-5.1959798994974875, 0.0)
  (-5.115577889447236, 0.0)
  (-5.035175879396985, 0.0)
  (-4.954773869346734, 0.0)
  (-4.874371859296482, 0.0)
  (-4.793969849246231, 0.0)
  (-4.71356783919598, 0.0)
  (-4.633165829145729, 0.0)
  (-4.552763819095477, 0.0)
  (-4.472361809045226, 0.0)
  (-4.391959798994975, 0.0)
  (-4.311557788944723, 0.0)
  (-4.231155778894473, 0.0)
  (-4.150753768844221, 0.0)
  (-4.0703517587939695, 0.0)
  (-3.9899497487437188, 0.0)
  (-3.909547738693467, 0.0)
  (-3.829145728643216, 0.0)
  (-3.748743718592965, 0.0)
  (-3.6683417085427137, 0.0)
  (-3.5879396984924625, 0.0)
  (-3.507537688442211, 0.0)
  (-3.4271356783919598, 0.0)
  (-3.3467336683417086, 0.0)
  (-3.2663316582914574, 0.0)
  (-3.185929648241206, 0.0)
  (-3.1055276381909547, 0.0)
  (-3.0251256281407035, 0.0)
  (-2.9447236180904524, 0.018425460636515883)
  (-2.864321608040201, 0.04522613065326627)
  (-2.7839195979899496, 0.07202680067001681)
  (-2.7035175879396984, 0.09882747068676719)
  (-2.6231155778894473, 0.12562814070351758)
  (-2.542713567839196, 0.15242881072026795)
  (-2.4623115577889445, 0.1792294807370185)
  (-2.3819095477386933, 0.20603015075376888)
  (-2.301507537688442, 0.23283082077051928)
  (-2.221105527638191, 0.25963149078726966)
  (-2.14070351758794, 0.28643216080402006)
  (-2.0603015075376883, 0.3132328308207706)
  (-1.979899497487437, 0.340033500837521)
  (-1.899497487437186, 0.36683417085427134)
  (-1.8190954773869348, 0.39363484087102174)
  (-1.7386934673366834, 0.4204355108877722)
  (-1.6582914572864322, 0.4472361809045226)
  (-1.5778894472361809, 0.47403685092127307)
  (-1.4974874371859297, 0.5008375209380235)
  (-1.4170854271356783, 0.5276381909547739)
  (-1.3366834170854272, 0.5544388609715243)
  (-1.256281407035176, 0.5812395309882746)
  (-1.1758793969849246, 0.6080402010050251)
  (-1.0954773869346734, 0.6348408710217756)
  (-1.015075376884422, 0.661641541038526)
  (-0.9346733668341709, 0.6884422110552763)
  (-0.8542713567839196, 0.7152428810720268)
  (-0.7738693467336684, 0.7420435510887772)
  (-0.6934673366834171, 0.7688442211055276)
  (-0.6130653266331658, 0.7956448911222781)
  (-0.5326633165829145, 0.8224455611390286)
  (-0.45226130653266333, 0.8492462311557789)
  (-0.37185929648241206, 0.8760469011725293)
  (-0.2914572864321608, 0.9028475711892797)
  (-0.21105527638190955, 0.9296482412060302)
  (-0.1306532663316583, 0.9564489112227806)
  (-0.05025125628140704, 0.983249581239531)
  (0.03015075376884422, 1.0100502512562815)
  (0.11055276381909548, 1.0368509212730317)
  (0.19095477386934673, 1.0636515912897824)
  (0.271356783919598, 1.0904522613065326)
  (0.35175879396984927, 1.117252931323283)
  (0.4321608040201005, 1.1440536013400335)
  (0.5125628140703518, 1.1708542713567838)
  (0.592964824120603, 1.1976549413735345)
  (0.6733668341708543, 1.2244556113902847)
  (0.7537688442211056, 1.2512562814070352)
  (0.8341708542713567, 1.2780569514237856)
  (0.914572864321608, 1.3048576214405359)
  (0.9949748743718593, 1.3316582914572865)
  (1.0753768844221105, 1.358458961474037)
  (1.1557788944723617, 1.3852596314907872)
  (1.236180904522613, 1.4120603015075377)
  (1.3165829145728642, 1.438860971524288)
  (1.3969849246231156, 1.4656616415410386)
  (1.4773869346733668, 1.492462311557789)
  (1.5577889447236182, 1.5192629815745393)
  (1.6381909547738693, 1.5460636515912898)
  (1.7185929648241205, 1.57286432160804)
  (1.7989949748743719, 1.5996649916247907)
  (1.879396984924623, 1.6264656616415412)
  (1.9597989949748744, 1.6532663316582914)
  (2.040201005025126, 1.6800670016750419)
  (2.120603015075377, 1.706867671691792)
  (2.201005025125628, 1.7336683417085428)
  (2.2814070351758793, 1.7604690117252932)
  (2.3618090452261304, 1.7872696817420435)
  (2.442211055276382, 1.814070351758794)
  (2.522613065326633, 1.8408710217755442)
  (2.6030150753768844, 1.8676716917922949)
  (2.6834170854271355, 1.8944723618090453)
  (2.763819095477387, 1.9212730318257958)
  (2.8442211055276383, 1.948073701842546)
  (2.9246231155778895, 1.9748743718592963)
  (3.0050251256281406, Inf)
  (3.0854271356783918, Inf)
  (3.1658291457286434, Inf)
  (3.2462311557788945, Inf)
  (3.3266331658291457, Inf)
  (3.407035175879397, Inf)
  (3.487437185929648, Inf)
  (3.5678391959798996, Inf)
  (3.648241206030151, Inf)
  (3.728643216080402, Inf)
  (3.809045226130653, Inf)
  (3.8894472361809047, Inf)
  (3.969849246231156, Inf)
  (4.050251256281407, Inf)
  (4.130653266331659, Inf)
  (4.211055276381909, Inf)
  (4.291457286432161, Inf)
  (4.371859296482412, Inf)
  (4.452261306532663, Inf)
  (4.532663316582915, Inf)
  (4.613065326633166, Inf)
  (4.693467336683417, Inf)
  (4.773869346733668, Inf)
  (4.8542713567839195, Inf)
  (4.934673366834171, Inf)
  (5.015075376884422, Inf)
  (5.0954773869346734, Inf)
  (5.175879396984925, Inf)
  (5.256281407035176, Inf)
  (5.336683417085427, Inf)
  (5.417085427135678, Inf)
  (5.49748743718593, Inf)
  (5.577889447236181, Inf)
  (5.658291457286432, Inf)
  (5.738693467336684, Inf)
  (5.819095477386934, Inf)
  (5.899497487437186, Inf)
  (5.9798994974874375, Inf)
  (6.060301507537688, Inf)
  (6.14070351758794, Inf)
  (6.221105527638191, Inf)
  (6.301507537688442, Inf)
  (6.381909547738694, Inf)
  (6.4623115577889445, Inf)
  (6.542713567839196, Inf)
  (6.623115577889447, Inf)
  (6.703517587939698, Inf)
  (6.78391959798995, Inf)
  (6.864321608040201, Inf)
  (6.944723618090452, Inf)
  (7.025125628140704, Inf)
  (7.105527638190955, Inf)
  (7.185929648241206, Inf)
  (7.266331658291457, Inf)
  (7.346733668341709, Inf)
  (7.42713567839196, Inf)
  (7.507537688442211, Inf)
  (7.5879396984924625, Inf)
  (7.668341708542713, Inf)
  (7.748743718592965, Inf)
  (7.8291457286432165, Inf)
  (7.909547738693467, Inf)
  (7.989949748743719, Inf)
  (8.07035175879397, Inf)
  (8.150753768844222, Inf)
  (8.231155778894472, Inf)
  (8.311557788944723, Inf)
  (8.391959798994975, Inf)
  (8.472361809045227, Inf)
  (8.552763819095478, Inf)
  (8.633165829145728, Inf)
  (8.71356783919598, Inf)
  (8.793969849246231, Inf)
  (8.874371859296483, Inf)
  (8.954773869346734, Inf)
  (9.035175879396984, Inf)
  (9.115577889447236, Inf)
  (9.195979899497488, Inf)
  (9.27638190954774, Inf)
  (9.35678391959799, Inf)
  (9.43718592964824, Inf)
  (9.517587939698492, Inf)
  (9.597989949748744, Inf)
  (9.678391959798995, Inf)
  (9.758793969849247, Inf)
  (9.839195979899497, Inf)
  (9.919597989949748, Inf)
  (10.0, Inf)
};
\addlegendentry{{}{minimize ($T=-3,L=3,P=2$)}}

\addplot+[
  mark = {none}
] coordinates {
  (-20.0, Inf)
  (-19.798994974874372, Inf)
  (-19.597989949748744, Inf)
  (-19.396984924623116, Inf)
  (-19.195979899497488, Inf)
  (-18.99497487437186, Inf)
  (-18.79396984924623, Inf)
  (-18.592964824120603, Inf)
  (-18.391959798994975, Inf)
  (-18.190954773869347, Inf)
  (-17.98994974874372, Inf)
  (-17.78894472361809, Inf)
  (-17.587939698492463, Inf)
  (-17.386934673366834, Inf)
  (-17.185929648241206, Inf)
  (-16.984924623115578, Inf)
  (-16.78391959798995, Inf)
  (-16.582914572864322, Inf)
  (-16.381909547738694, Inf)
  (-16.180904522613066, Inf)
  (-15.979899497487438, Inf)
  (-15.77889447236181, Inf)
  (-15.577889447236181, Inf)
  (-15.376884422110553, Inf)
  (-15.175879396984925, Inf)
  (-14.974874371859297, Inf)
  (-14.773869346733669, Inf)
  (-14.57286432160804, Inf)
  (-14.371859296482413, Inf)
  (-14.170854271356784, Inf)
  (-13.969849246231156, Inf)
  (-13.768844221105528, Inf)
  (-13.5678391959799, Inf)
  (-13.366834170854272, Inf)
  (-13.165829145728644, Inf)
  (-12.964824120603016, Inf)
  (-12.763819095477388, Inf)
  (-12.56281407035176, Inf)
  (-12.361809045226131, Inf)
  (-12.160804020100503, Inf)
  (-11.959798994974875, Inf)
  (-11.758793969849247, Inf)
  (-11.557788944723619, Inf)
  (-11.35678391959799, Inf)
  (-11.155778894472363, Inf)
  (-10.954773869346734, Inf)
  (-10.753768844221106, Inf)
  (-10.552763819095478, Inf)
  (-10.35175879396985, Inf)
  (-10.150753768844222, Inf)
  (-9.949748743718592, Inf)
  (-9.748743718592964, Inf)
  (-9.547738693467336, Inf)
  (-9.346733668341708, Inf)
  (-9.14572864321608, Inf)
  (-8.944723618090451, Inf)
  (-8.743718592964823, Inf)
  (-8.542713567839195, Inf)
  (-8.341708542713567, Inf)
  (-8.140703517587939, Inf)
  (-7.939698492462312, Inf)
  (-7.738693467336684, Inf)
  (-7.5376884422110555, Inf)
  (-7.336683417085427, Inf)
  (-7.135678391959799, Inf)
  (-6.934673366834171, Inf)
  (-6.733668341708543, Inf)
  (-6.532663316582915, Inf)
  (-6.331658291457287, Inf)
  (-6.130653266331659, Inf)
  (-5.9296482412060305, Inf)
  (-5.728643216080402, Inf)
  (-5.527638190954774, Inf)
  (-5.326633165829146, Inf)
  (-5.125628140703518, Inf)
  (-4.924623115577889, Inf)
  (-4.723618090452261, Inf)
  (-4.522613065326633, Inf)
  (-4.321608040201005, Inf)
  (-4.1206030150753765, Inf)
  (-3.919597989949749, Inf)
  (-3.7185929648241207, Inf)
  (-3.5175879396984926, Inf)
  (-3.3165829145728645, Inf)
  (-3.1155778894472363, Inf)
  (-2.9145728643216082, 0.4946608040201005)
  (-2.71356783919598, 0.48209798994974873)
  (-2.512562814070352, 0.46953517587939697)
  (-2.3115577889447234, 0.4569723618090452)
  (-2.1105527638190953, 0.44440954773869346)
  (-1.9095477386934674, 0.4318467336683417)
  (-1.7085427135678393, 0.41928391959798994)
  (-1.5075376884422111, 0.4067211055276382)
  (-1.306532663316583, 0.3941582914572864)
  (-1.1055276381909547, 0.38159547738693467)
  (-0.9045226130653267, 0.3690326633165829)
  (-0.7035175879396985, 0.35646984924623115)
  (-0.5025125628140703, 0.3439070351758794)
  (-0.3015075376884422, 0.33134422110552764)
  (-0.10050251256281408, 0.3187814070351759)
  (0.10050251256281408, 0.3062185929648241)
  (0.3015075376884422, 0.29365577889447236)
  (0.5025125628140703, 0.2810929648241206)
  (0.7035175879396985, 0.26853015075376885)
  (0.9045226130653267, 0.2559673366834171)
  (1.1055276381909547, 0.24340452261306533)
  (1.306532663316583, 0.23084170854271358)
  (1.5075376884422111, 0.21827889447236182)
  (1.7085427135678393, 0.20571608040201006)
  (1.9095477386934674, 0.1931532663316583)
  (2.1105527638190953, 0.18059045226130654)
  (2.3115577889447234, 0.1680276381909548)
  (2.512562814070352, 0.155464824120603)
  (2.71356783919598, 0.14290201005025124)
  (2.9145728643216082, 0.13033919597989949)
  (3.1155778894472363, 0.11777638190954773)
  (3.3165829145728645, 0.10521356783919597)
  (3.5175879396984926, 0.09265075376884421)
  (3.7185929648241207, 0.08008793969849246)
  (3.919597989949749, 0.0675251256281407)
  (4.1206030150753765, 0.05496231155778897)
  (4.321608040201005, 0.04239949748743721)
  (4.522613065326633, 0.029836683417085452)
  (4.723618090452261, 0.017273869346733695)
  (4.924623115577889, 0.004711055276381937)
  (5.125628140703518, 0.0)
  (5.326633165829146, 0.0)
  (5.527638190954774, 0.0)
  (5.728643216080402, 0.0)
  (5.9296482412060305, 0.0)
  (6.130653266331659, 0.0)
  (6.331658291457287, 0.0)
  (6.532663316582915, 0.0)
  (6.733668341708543, 0.0)
  (6.934673366834171, 0.0)
  (7.135678391959799, 0.0)
  (7.336683417085427, 0.0)
  (7.5376884422110555, 0.0)
  (7.738693467336684, 0.0)
  (7.939698492462312, 0.0)
  (8.140703517587939, 0.0)
  (8.341708542713567, 0.0)
  (8.542713567839195, 0.0)
  (8.743718592964823, 0.0)
  (8.944723618090451, 0.0)
  (9.14572864321608, 0.0)
  (9.346733668341708, 0.0)
  (9.547738693467336, 0.0)
  (9.748743718592964, 0.0)
  (9.949748743718592, 0.0)
  (10.150753768844222, 0.0)
  (10.35175879396985, 0.0)
  (10.552763819095478, 0.0)
  (10.753768844221106, 0.0)
  (10.954773869346734, 0.0)
  (11.155778894472363, 0.0)
  (11.35678391959799, 0.0)
  (11.557788944723619, 0.0)
  (11.758793969849247, 0.0)
  (11.959798994974875, 0.0)
  (12.160804020100503, 0.0)
  (12.361809045226131, 0.0)
  (12.56281407035176, 0.0)
  (12.763819095477388, 0.0)
  (12.964824120603016, 0.0)
  (13.165829145728644, 0.0)
  (13.366834170854272, 0.0)
  (13.5678391959799, 0.0)
  (13.768844221105528, 0.0)
  (13.969849246231156, 0.0)
  (14.170854271356784, 0.0)
  (14.371859296482413, 0.0)
  (14.57286432160804, 0.0)
  (14.773869346733669, 0.0)
  (14.974874371859297, 0.0)
  (15.175879396984925, 0.0)
  (15.376884422110553, 0.0)
  (15.577889447236181, 0.0)
  (15.77889447236181, 0.0)
  (15.979899497487438, 0.0)
  (16.180904522613066, 0.0)
  (16.381909547738694, 0.0)
  (16.582914572864322, 0.0)
  (16.78391959798995, 0.0)
  (16.984924623115578, 0.0)
  (17.185929648241206, 0.0)
  (17.386934673366834, 0.0)
  (17.587939698492463, 0.0)
  (17.78894472361809, 0.0)
  (17.98994974874372, 0.0)
  (18.190954773869347, 0.0)
  (18.391959798994975, 0.0)
  (18.592964824120603, 0.0)
  (18.79396984924623, 0.0)
  (18.99497487437186, 0.0)
  (19.195979899497488, 0.0)
  (19.396984924623116, 0.0)
  (19.597989949748744, 0.0)
  (19.798994974874372, 0.0)
  (20.0, 0.0)
};
\addlegendentry{{}{maximize ($T=5,L=-3,P=0.5$)}}

\end{axis}
\end{tikzpicture}